\documentclass[letterpaper, 10 pt, conference]{ieeeconf}  

\IEEEoverridecommandlockouts                              

\usepackage{graphics} 
\usepackage{epsfig} 
\usepackage{mathptmx} 
\usepackage{times} 
\usepackage{amsmath} 
\usepackage{amssymb}  
\usepackage[noadjust]{cite}

\newtheorem{theorem}{Theorem}
\newtheorem{lemma}{Lemma}
\newtheorem{remark}{Remark}
\newtheorem{corollary}{Corollary}[theorem]

\DeclareRobustCommand{\svdots}{
  \vcenter{%
    \offinterlineskip
    \hbox{.}
    \vskip0.25\normalbaselineskip
    \hbox{.}
    \vskip0.25\normalbaselineskip
    \hbox{.}%
  }%
}

\title{\LARGE \bf
Least Squares Training of Quadratic Convolutional Neural Networks with Applications to System Theory
}

\author{Zachary Yetman Van Egmond$^{1}$ and Luis Rodrigues$^{1}$
\thanks{$^{1}$Zachary Yetman Van Egmond and Luis Rodrigues are with the Department of Electrical and Computer Engineering, Concordia University, Montreal, QC H3G~2W1, Canada}%
\thanks{{\tt\small zachary.yetmanvanegmond@mail.concordia.ca}}%
\thanks{{\tt\small luis.rodrigues@concordia.ca}}%
}

\begin{document}

\maketitle
\thispagestyle{empty}
\pagestyle{empty}

\begin{abstract}
This paper provides a least squares formulation for the training of a 2-layer convolutional neural network using quadratic activation functions, a 2-norm loss function, and no regularization term. 
Using this method, an analytic expression for the globally optimal weights is obtained alongside a quadratic input-output equation for the network. 
These properties make the network a viable tool in system theory by enabling further analysis, such as the sensitivity of the output to perturbations in the input, which is crucial for safety-critical systems such as aircraft or autonomous vehicles.
The least squares method is compared to previously proposed strategies for training quadratic networks and to a back-propagation-trained ReLU network. 
The proposed method is applied to a system identification problem and a GPS position estimation problem. 
The least squares network is shown to have a significantly reduced training time with minimal compromises on prediction accuracy alongside the advantages of having an analytic input-output equation. 
Although these results only apply to 2-layer networks, this paper motivates the exploration of deeper quadratic networks in the context of system theory.

\end{abstract}

\section{INTRODUCTION}
Neural networks have shown great potential for classification and regression problems in various fields. However, despite this success, neural networks present several shortcomings that limit their applicability to certain engineering problems. One of the primary limitations is the inability to make concrete statements about the behavior and predictions of the network. This issue is of particular importance when considering their use in safety-critical systems \cite{safetyCritNN_2020}, which may require guarantees for certification. Other limitations include the excessive training times caused by back-propagation (BP) methods and the presence of large numbers of hyper-parameters that must be tuned.

One of the types of neural networks that aims to address these issues is quadratic neural networks (QNNs), which use a second-order polynomial as their activation function. In 2021, the authors of \cite{bartan2021neural,bartanThesis_2022} showed that the training of a 2-layer QNN with constraints placed on its weights and regularization is a convex problem with guarantees of finding the globally optimal weights. Furthermore, their formulation allowed for the input-output equation of the network to be written in a quadratic form. Extending this work, the author of \cite{luis2023leastsquares} showed that without regularization the training of a QNN could be formulated as a least squares problem. This allowed for the analytic training of QNNs which drastically reduced training times. Further work on QNNs presented in \cite{qnnControl_2023} demonstrated their ability to be used for system identification and controller design with Lyapunov-based stability guarantees.

This paper extends the work of \cite{bartan2021neural, bartanThesis_2022, luis2023leastsquares} by formulating a least squares solution to the training of a 2-layer quadratic convolutional neural network (CQNN). To achieve this, the CQNN problem is first transformed into an equivalent QNN problem, which drastically reduces the number of learned weights compared to the CQNN formulation given in \cite{bartan2021neural}. The resulting convolutional network has both an analytic expression for its globally optimal weights and a smooth input-output mapping.

The use of least squares for analytical neural network training has been of interest for some time. An early implementation presented in \cite{pseudoinverse_2001} outlined training each layer of a neural network as a least squares problem which was minimized by relating each layer's weights to the pseudoinverse of the layer's inputs. Though the resulting network had perfect training accuracy, the total number of neurons in each layer was equal to the number of training samples.
A similar layer-by-layer training method was presented by the authors of \cite{analyticMLP_2021}, which trained the network in two-layer modules by generating pseudo-labels by a random linear transformation of the final labels. This idea was expanded upon to allow for the training of convolutional neural networks (CNNs) by the authors of \cite{analyticCNN_2024}. However, \cite{analyticCNN_2024} is focused on pattern recognition, and the extension of the work to system theory is not obvious due to the lack of an explicit closed-form input-output expression for the network. Furthermore, due to the method of pseudo-label generation the resulting solution is not guaranteed to be the global minimum.

This paper is divided as follows. Section~\ref{sec:prelim} presents the required preliminaries. Section~\ref{sec:CQNNasQNN} shows the formulation of a CQNN as an equivalent QNN problem. Section~\ref{sec:CQNNasLS} proposes a formulation for training a CQNN using least squares. Section~\ref{sec:results} applies the least squares CQNN to several examples.

\section{PRELIMINARIES}
\label{sec:prelim}
\subsection{Quadratic Neural Networks}
The input-output expression for a 2-layer, single-output QNN, with $m$ neurons in the hidden layer is given as
\begin{align}
    \label{eqn_NNInOut}
    \hat{y}(x) = \sum_{j=1}^{m} \alpha_j \sigma(w_j^T x)
\end{align}
where $\sigma(z)$ is a quadratic activation function defined as
\begin{equation}
    \label{eqn_quadactivation}
    \sigma(z)=az^2+bz+c
\end{equation}
with $a\neq0,b,c$ being constants that parametrize the quadratic activation function. This paper will use the following notation. Using the sample index $i$, $\hat{y}_i\in\mathbb{R}$ and $y_i\in\mathbb{R}$ correspond to the network output and desired output (or label) for a given input vector $x_i\in\mathbb{R}^{n}$ containing $n$ features. The first and second layer weights of the network for the $j^{th}$ neuron are given by $w_j\in\mathbb{R}^{n}$ and $\alpha_j\in\mathbb{R}$ respectively.
The training of a QNN defined by input-output equation (\ref{eqn_NNInOut}) can be written as a non-convex optimization problem \cite{bartan2021neural}.
\begin{equation}
    \label{eqn_NNOptim}
    \begin{split}
        \min\; & l(\hat{y},y) + \beta \sum_{j=1}^m ||\alpha_j||_1 \\
    \text{s.t.}\; & \hat{y}_i = \sum_{j=1}^{m} \alpha_j\sigma(w_j^Tx_i),\; i=1,\dots,N \\
    & ||w_j||_2 = 1, \; j=1,\dots,m
    \end{split}
\end{equation}
where $l(\cdot)$ is a convex loss function which is applied to all network outputs and labels and $\beta\geq0$ is a regularization term.

\begin{lemma}
    \cite{bartan2021neural} For fixed $a\neq0$, $b$, $c$, regularization term $\beta\geq0$, and convex loss function $l(\cdot)$, the non-convex problem (\ref{eqn_NNOptim}) has the same global optimal solution as the convex problem given by
    \begin{equation}
        \label{eqn_QNNConvex}
        \begin{split}
        \min\; & l(\hat{y},y)+\beta (Z^{4}_{+} + Z^{4}_{-}) \\
        \text{s.t.}\; & \hat{y}_i = \begin{bmatrix} x_i \\ 1 \end{bmatrix}^T \begin{bmatrix} aZ^{1} & \frac{b}{2}Z^{2} \\ \frac{b}{2}(Z^{2})^T & cZ^{4} \end{bmatrix} \begin{bmatrix} x_i \\ 1 \end{bmatrix} \\
        & Z^{q} = Z_{+}^{q} - Z_{-}^{q},\;q=1,2,4 \\
        & Z_{+}^{4} = \textbf{Trace}(Z_{+}^{1}), \; Z_{-}^{4} = \textbf{Trace}(Z_{-}^{1}) \\
        & Z_{+} = \begin{bmatrix} Z_{+}^{1} & Z_{+}^{2} \\ (Z_{+}^{2})^T & Z_{+}^{4} \end{bmatrix}, \; Z_{-} = \begin{bmatrix} Z_{-}^{1} & Z_{-}^{2} \\ (Z_{-}^{2})^T & Z_{-}^{4} \end{bmatrix} \\
        & Z_{+} \geq 0, \; Z_{-} \geq 0 \\
        & i=1,\dots,N
        \end{split}
    \end{equation}
    when the number of neurons $m\geq m^*$ where
    \begin{align}
        m^* = \text{rank}(Z_{+}^*) + \text{rank}(Z_{-}^*)
    \end{align}
    with $Z_{+}^*,Z_{-}^*\in\mathbb{R}^{(n+1)\times(n+1)}$ being the solutions to the optimization problem (\ref{eqn_QNNConvex}).
\end{lemma}

\begin{lemma}
    \label{lem:QNNnoLMIconst}
    \cite{luis2023leastsquares} If $\beta=0$, $a>0$, $c>0$, and $b^2-4ac \geq 0$ then any solution of
    \begin{equation}
        \label{eqn_QNNnoLMIconst}
        \begin{split}
        \min\; & l(\hat{y},y) \\
        \text{s.t.}\; & \hat{y}_i = \begin{bmatrix} x_i \\ 1 \end{bmatrix}^T \begin{bmatrix} aZ^{1} & \frac{b}{2}Z^{2} \\ \frac{b}{2}(Z^{2})^T & cZ^{4} \end{bmatrix} \begin{bmatrix} x_i \\ 1 \end{bmatrix}  \\
        & Z^{4} = \textbf{Trace}(Z^{1})
        \end{split}
    \end{equation}
    is also a solution of (\ref{eqn_QNNConvex}) where $Z^q=Z^q_+-Z^q_-$, $q=1,2,4$.
\end{lemma}
This result will be important in Section~\ref{sec:CQNNasQNN} where the training of a CQNN will be transformed to take the form (\ref{eqn_QNNnoLMIconst}).

\subsection{Quadratic Convolutional Neural Networks}
Consider a vector of length $n$
\begin{align}
    x = \begin{bmatrix} x_1 & x_2 & \cdots & x_n\end{bmatrix}^T
\end{align}
from which $K$ patches of data are formed such that each patch $\chi_k$ contains $f$ sequential data points of $x$. Since the last element of $\chi_K$ is also the last element of $x$, it can be seen that $K = n-f+1$.
\begin{subequations}
\begin{align}
    \chi = \begin{bmatrix} \chi_1 & \chi_2 & \cdots & \chi_k & \cdots & \chi_K\end{bmatrix}^T \\
    \chi_k = \begin{bmatrix} x_k & x_{k+1} & \cdots  & x_{k+f-1}\end{bmatrix}^T
\end{align}
\end{subequations}
Using this data patch definition of $\chi_k$, (\ref{eqn_CNNInOut}) represents the input-output expression of a 2-layer CQNN using $\bar{m}$ filters of length $f$ and stride 1, and activation function $\sigma(z)$ defined by (\ref{eqn_quadactivation}).
\begin{equation}
    \label{eqn_CNNInOut}
    \hat{y}(x) = \sum_{j=1}^{\bar{m}} \sum_{k=1}^K \alpha_{jk}\sigma(w_j^T\chi_k)
\end{equation}
The training of the CQNN defined by input-output equation (\ref{eqn_CNNInOut}) can be written as a non-convex optimization problem \cite{bartan2021neural}.
\begin{equation}
    \label{eqn_CNNOptim}
    \begin{split}
    \min\; & l(\hat{y},y) + \beta \sum_{j=1}^{\bar{m}} ||\alpha_j||_1 \\
    \text{s.t.}\; & \hat{y}_i = \sum_{j=1}^{\bar{m}} \sum_{k=1}^K \alpha_{jk}\sigma(w_j^T\chi_{i,k}), \; i=1,\dots,N \\
    & ||w_j||_2 = 1, \; j=1,\dots,\bar{m}
    \end{split}
\end{equation}
where $\chi_{i,k}$ is patch $\chi_k$ of input sample $x_i$ and the second layer weights are given by $\alpha_j=[\alpha_{j1} \; \cdots \; \alpha_{jK}]^T$.

\begin{lemma}
    \cite{bartan2021neural} For fixed $a\neq0$, $b$, $c$, regularization term $\beta\geq0$, and convex loss function $l(\cdot)$, the non-convex problem (\ref{eqn_CNNOptim}) has the same global optimal solution as the convex problem given by
    \begin{equation}
        \label{eqn_CQNNConvex}
        \begin{split}
        \min\; & l(\hat{y},y)+\beta \sum_{k=1}^K(Z^{k,4}_{+} + Z^{k,4}_{-}) \\
        \text{s.t.}\; & \hat{y}_i = \sum_{k=1}^K \left( \begin{bmatrix} \chi_{i,k} \\ 1 \end{bmatrix}^T \begin{bmatrix} aZ^{k,1} & \frac{b}{2}Z^{k,2} \\ \frac{b}{2}(Z^{k,2})^T & cZ^{k,4} \end{bmatrix} \begin{bmatrix} \chi_{i,k} \\ 1 \end{bmatrix} \right)  \\
        & Z^{k,q} = Z_{+}^{k,q} - Z_{-}^{k,q},\;q=1,2,4  \\
        & Z_{+}^{k,4} = \textbf{Trace}(Z_{+}^{k,1}), \; Z_{-}^{k,4} = \textbf{Trace}(Z_{-}^{k,1})  \\
        & Z_{+}^{k} = \begin{bmatrix} Z_{+}^{k,1} & Z_{+}^{k,2} \\ (Z_{+}^{k,2})^T & Z_{+}^{k,4} \end{bmatrix}, \; Z_{-}^{k} = \begin{bmatrix} Z_{-}^{k,1} & Z_{-}^{k,2} \\ (Z_{-}^{k,2})^T & Z_{-}^{k,4} \end{bmatrix}  \\
        & Z_{+}^{k} \geq 0, \; Z_{-}^{k} \geq 0  \\
        & k=1 ,\dots, K,\quad i=1 ,\dots, N 
        \end{split}
    \end{equation}
    when the number of filters $\bar{m}\geq \bar{m}^*$ where
    \begin{align}
        \bar{m}^* = K\sum_{k=1}^K\left(\text{rank}(Z^{k*}_{+}) + \text{rank}(Z^{k*}_{-})\right)
    \end{align}
    with $Z^{k*}_{+},Z^{k*}_{-}\in\mathbb{R}^{(f+1)\times(f+1)}$ for $k=1,\dots,K$ being the solutions to the optimization problem (\ref{eqn_CQNNConvex}).
\end{lemma}

\section{CQNN Formulated as a QNN}
\label{sec:CQNNasQNN}
The following lemma is an extension of Lemma~\ref{lem:QNNnoLMIconst} which closely follows the proof from \cite{luis2023leastsquares} and will be used in the proof of the main theorem of this paper.

\begin{figure}[t]
    \centering
    \includegraphics[width=0.48\textwidth]{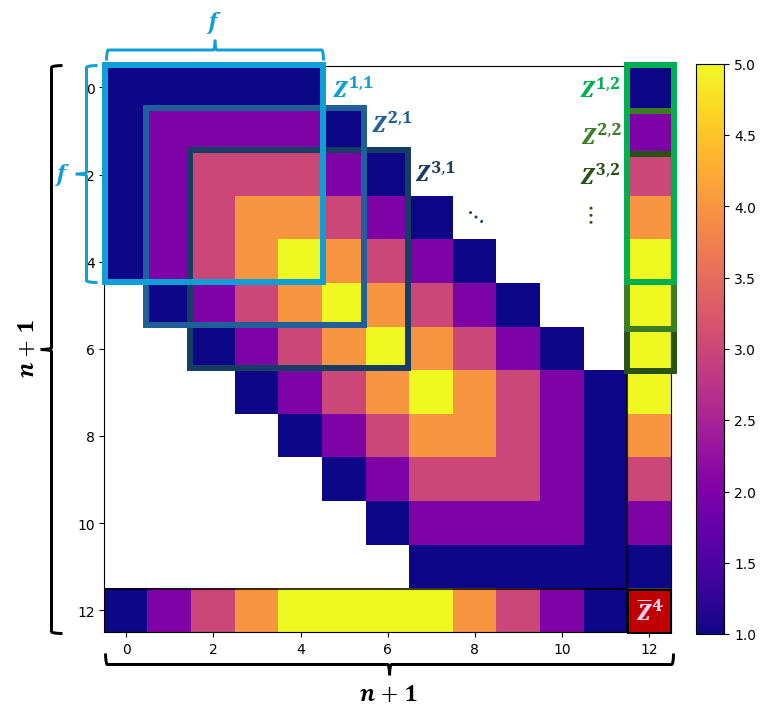} 
    \caption{Visualization of $\bar{Z}^1$ and $\bar{Z}^2$, for $f=5$, stride 1, and $n=12$. White squares represent zero entries, colored squares show the number of overlaps of different $Z^{k,1}$ and $Z^{k,2}$ to form the elements of $\bar{Z}^1$ and $\bar{Z}^2$.} 
    \label{fig:CQNN}
\end{figure}

\begin{lemma}
    If $\beta=0$, $a>0$, $c>0$, and $b^2-4ac \geq 0$ then any solution of
    \begin{equation}
        \label{eqn_CQNNnoLMIconst}
        \begin{split}
        \min\; & l(\hat{y},y) \\
        \text{s.t.}\; & \hat{y}_i = \sum_{k=1}^K \left( \begin{bmatrix} \chi_{i,k} \\ 1 \end{bmatrix}^T \begin{bmatrix} aZ^{k,1} & \frac{b}{2}Z^{k,2} \\ \frac{b}{2}(Z^{k,2})^T & cZ^{k,4} \end{bmatrix} \begin{bmatrix} \chi_{i,k} \\ 1 \end{bmatrix} \right) \\
        & Z^{k,4} = \textbf{Trace}(Z^{k,1})
        \end{split}
    \end{equation}
    is also a solution of (\ref{eqn_CQNNConvex}) where $Z^{k,q}=Z^{k,q}_+-Z^{k,q}_-$, $q=1,2,4$. 
\end{lemma}
\begin{proof}
    The proof is the same for all samples, so the index $i$ is dropped. Let there be a solution to (\ref{eqn_CQNNConvex}) where $a>0$, $b\neq0$, $c>0$, consider the matrix for the $k^{th}$ filter
    \begin{align*}
        Z^k=\begin{bmatrix}
            aZ^{k,1} & \frac{b}{2}Z^{k,2} \\ \frac{b}{2}(Z^{k,2})^T & c\textbf{Trace}(Z^{k,1})
        \end{bmatrix}
    \end{align*}
    Since $Z^{k}$ is symmetric it can be written as the difference of two positive semidefinite matrices $Z^k=Z^k_+-Z^k_-$ where
    \begin{align*}
        Z^k_{+/-} = \begin{bmatrix}
            aZ^{k,1}_{+/-} & \frac{b}{2}Z^{k,2}_{+/-} \\ \frac{b}{2}(Z^{k,2}_{+/-})^T & c\textbf{Trace}(Z^{k,1}_{+/-})
        \end{bmatrix}
    \end{align*}
    Using Schur's complement, the positive definite constraints are equivalent to
    \begin{align*}
        c\textbf{Trace}(Z^{k,1}_+) &\geq 0\\
        \left(1-\textbf{Trace}(Z^{k,1}_+)\textbf{Trace}^{\dag}(Z^{k,1}_+)\right)\frac{b}{2}\left(Z^{k,2}_+\right)^T &= 0 \\
        aZ^{k,1}_+-\frac{b^2}{4}Z^{k,2}_+\textbf{Trace}^{\dag}(cZ^{k,1}_+)(Z^{k,2}_+)^T &\geq0
    \end{align*}
    where, defining $t^+=\textbf{Trace}(cZ^{k,1}_+)$
    \begin{align*}
        \textbf{Trace}^{\dag}(cZ^{k,1}_+) = \begin{cases}
            0 & t^+=0 \\ \textbf{Trace}^{-1}(cZ^{k,1}_+) & t^+\neq0
        \end{cases}
    \end{align*}
    is the Moore-Penrose pseudo-inverse of $\textbf{Trace}(cZ^{k,1}_+)$. When $t^+=0$, the conditions become $Z^{k,2}_+=0$, $Z^{k,1}\geq0$, which satisfy the matrix inequalities of (\ref{eqn_CQNNConvex}). When $t^+\neq0$, the condition becomes
    \begin{align*}
        Z^{k,2}_+(Z^{k,2}_+)^T \leq \frac{4ac}{b^2}Z^{k,1}_{+}\textbf{Trace}(Z^{k,1}_+)
    \end{align*}
    which by Schur's complement satisfies the matrix inequalities of (\ref{eqn_CQNNConvex}) if $b^2-4ac\geq0$ and $ac>0$. The same applies for $Z^{k,1}_-$ and $Z^{k,2}_-$. Furthermore, since this holds regardless of $k$, a solution to (\ref{eqn_CQNNnoLMIconst}) is also a solution to (\ref{eqn_CQNNConvex}).
\end{proof}

\begin{figure}[t]
    \centering
    \includegraphics[width=0.45\textwidth]{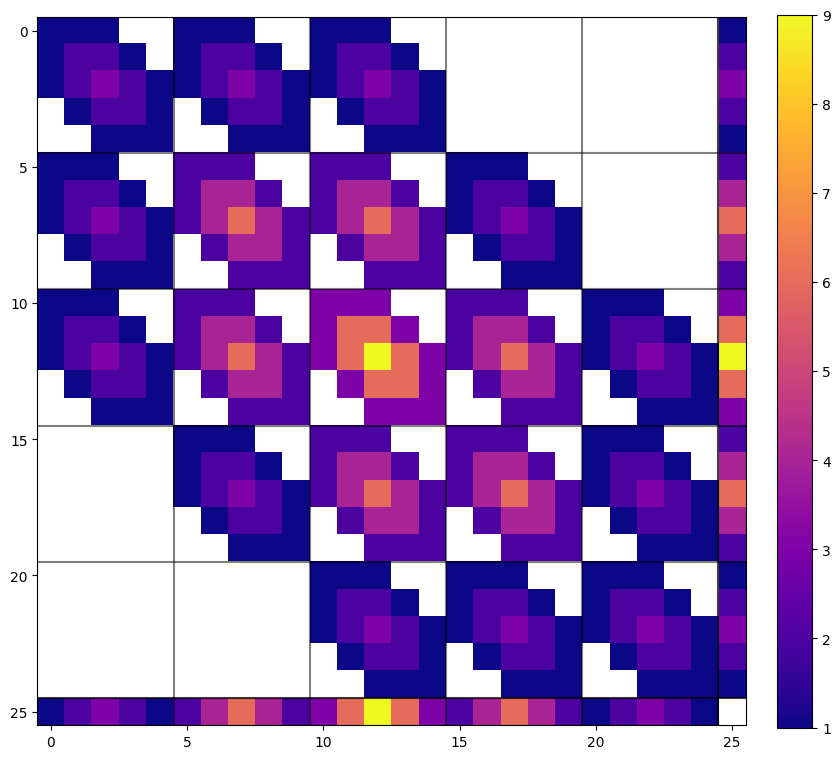}
    \caption{Visualization of $\bar{Z}^1$ and $\bar{Z}^2$ for a 2D convolution with $3\times 3$ filter on a $5\times5$ piece of 2D data. White squares represents zero entries, colored squares show the number of overlaps of different $Z^{k,1}$ and $Z^{k,2}$ to form the elements of $\bar{Z}^1$ and $\bar{Z}^2$.} 
    \label{fig:2d_convolution}
\end{figure}

\begin{theorem}
    \label{theo:CQNNasQNN}
    The training of a CQNN with filter size $f$ and stride 1 of the form (\ref{eqn_CQNNnoLMIconst}) can be formulated as a QNN training problem of the form (\ref{eqn_QNNnoLMIconst}) with additional constraints imposed on the structure of its $Z$ matrices as follows
    \begin{equation}
        \label{eqn_cqnntheorem}
        \begin{split}
        \min\; & l(\hat{y},y) \\
        \text{s.t.}\; & \hat{y}_i = \begin{bmatrix} x_{i} \\ 1 \end{bmatrix}^T \begin{bmatrix} a\bar{Z}^{1} & \frac{b}{2}\bar{Z}^{2} \\ \frac{b}{2}(\bar{Z}^{2})^T & c\bar{Z}^{4} \end{bmatrix} \begin{bmatrix} x_{i} \\ 1 \end{bmatrix} \\
        & \bar{Z}^{4} = \textbf{Trace}(\bar{Z}^{1}) \\
        & \bar{Z}^1_{r,c} = 0 \quad \text{if} \quad |r-c| \geq f
        \end{split}
    \end{equation}
    where $\bar{Z}^1_{r,c}$ is the entry at the $r^{th}$ row and $c^{th}$ column of $\bar{Z}^1$.
\end{theorem}
\begin{proof}
    For this proof, the sample index subscript $i$ will be dropped for convenience. Let
    \begin{align*}
        x_{n:m} &= \begin{bmatrix} x_n & x_{n+1} & \cdots & x_m \end{bmatrix}^T \\
        x &= \begin{bmatrix} x_{1:k-1}^T & \chi_k^T & x_{k+f:n}^T \end{bmatrix}^T
    \end{align*}
    Starting from the input-output equation given in (\ref{eqn_CQNNnoLMIconst}), additional terms are added to the $[\chi_k^T\;1]^T$ vector in such a way that it becomes $[x^T\;1]^T$.
    \begin{align*}
        \hat{y} = \sum_{k=1}^K \left( \begin{bmatrix} \chi_{k} \\ 1 \end{bmatrix}^T \begin{bmatrix} aZ^{k,1} & \frac{b}{2}Z^{k,2} \\ \frac{b}{2}(Z^{k,2})^T & cZ^{k,4} \end{bmatrix} \begin{bmatrix} \chi_{k} \\ 1 \end{bmatrix} \right) =\\
        \sum_{k=1}^K \left( \begin{bmatrix} x_{1:k-1} \\ \chi_k \\ x_{k+f:n} \\ 1 \end{bmatrix}^T \begin{bmatrix} 0 & 0 & 0 & 0 \\ 0 & aZ^{k,1} & 0 & \frac{b}{2}Z^{k,2} \\ 0 & 0 & 0 & 0 \\ 0 & \frac{b}{2}(Z^{k,2}) & 0 & cZ^{k,4} \end{bmatrix} \begin{bmatrix} x_{1:k-1} \\ \chi_k \\ x_{k+f:n}^T \\ 1 \end{bmatrix} \right)
    \end{align*}
    The inner matrix is padded with zeros corresponding to the rows and columns which multiply the added vector entries (padding will vary depending on $k$). Since the vectors now contain all the elements of $x$, they are no longer dependent on $k$ and can be removed from the summation.
    \begin{align*}
        \hat{y} &= \sum_{k=1}^K \left( \begin{bmatrix} x \\ 1 \end{bmatrix}^T \begin{bmatrix} 0 & 0 & 0 & 0 \\ 0 & aZ^{k,1} & 0 & \frac{b}{2}Z^{k,2} \\ 0 & 0 & 0 & 0 \\ 0 & \frac{b}{2}(Z^{k,2}) & 0 & cZ^{k,4} \end{bmatrix} \begin{bmatrix} x \\ 1 \end{bmatrix} \right) \\
        &= \begin{bmatrix} x \\ 1 \end{bmatrix}^T \sum_{k=1}^K \left(\begin{bmatrix} 0 & 0 & 0 & 0 \\ 0 & aZ^{k,1} & 0 & \frac{b}{2}Z^{k,2} \\ 0 & 0 & 0 & 0 \\ 0 & \frac{b}{2}(Z^{k,2})^T & 0 & cZ^{k,4} \end{bmatrix}\right) \begin{bmatrix} x \\ 1 \end{bmatrix} \\
        &= \begin{bmatrix} x \\ 1 \end{bmatrix}^T \begin{bmatrix} a\bar{Z}^{1} & \frac{b}{2}\bar{Z}^{2} \\ \frac{b}{2}(\bar{Z}^{2})^T & c\bar{Z}^{4} \end{bmatrix} \begin{bmatrix} x \\ 1 \end{bmatrix}
    \end{align*}
    where
    \begin{equation}
        \label{eqn_defineZBar}
        \begin{split}
        \bar{Z}^1 &= \sum_{k=1}^K\begin{bmatrix} O_{k-1,k-1} & O_{k-1,f} & O_{k-1,p} \\ O_{f,k-1} & Z^{k,1} & O_{f,p} \\ O_{p,k-1} & O_{p,f} & O_{p,p} \end{bmatrix} \\
        \bar{Z}^2 &= \sum_{k=1}^K \begin{bmatrix} (O_{k-1,1})^T & (Z^{k,2})^T & (O_{p,1})^T \end{bmatrix}^T  \\
        \bar{Z}^4 &= \sum_{k=1}^K Z^{k,4}
        \end{split}
    \end{equation}
    with $p=n-k-f+1$ and where $O_{r,c}$ is an $r\times c$ matrix of zeros. By examination, $\bar{Z}^1_{r,c} = 0$ if $|r-c| \geq f$, where $\bar{Z}^1_{r,c}$ is the element at the $r^{th}$ row and $c^{th}$ column of $\bar{Z}^1$. This is best shown visually in Fig.~\ref{fig:CQNN}, where the non-zero entries form a strip down the central diagonal. Finally, the trace condition still holds as shown below.
    \begin{equation*}
        \begin{split}
        \textbf{Trace}(\bar{Z}^1) &= \textbf{Trace}\left(\sum_{k=1}^K\begin{bmatrix} O_{k-1,k-1} & O_{k-1,f} & O_{k-1,p} \\ O_{f,k-1} & Z^{k,1} & O_{f,p} \\ O_{p,k-1} & O_{p,f} & O_{p,p} \end{bmatrix}\right) \\
        &= \sum_{k=1}^K\textbf{Trace}(Z^{k,1}) = \sum_{k=1}^KZ^{k,4}=\bar{Z}^4
        \end{split}
    \end{equation*}
\end{proof}

\begin{figure}
    \centering
    \includegraphics[width=0.34\textwidth]{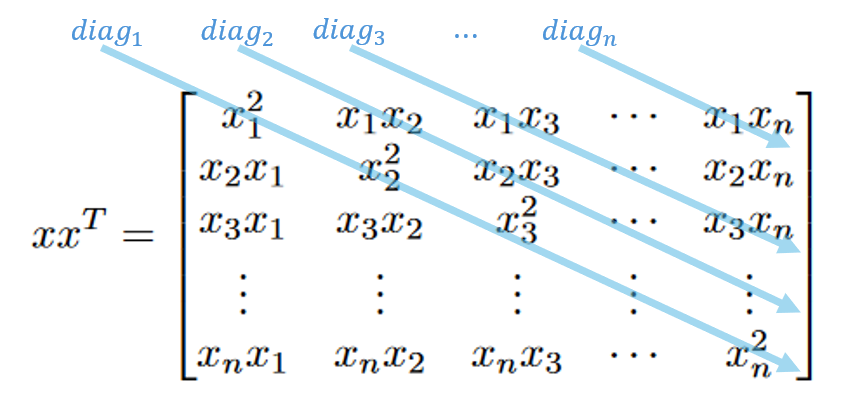}
    \caption{Entries of $xx^T$ with diagonals marked} 
    \label{fig:xxT_diags}
\end{figure}
\begin{remark}[Number of weights]
    Recalling that $Z^1$ is symmetric and $Z^4=\textbf{Trace}(Z^1)$, the solution to (\ref{eqn_CQNNnoLMIconst}) requires solving for $\frac{(f+3)(n-f+1)f}{2}$ total unique weights. Comparatively, the solution to (\ref{eqn_cqnntheorem}) requires $\frac{(2n-f+1)f}{2}+n$ total unique weights, which is always less given $f\leq n$ and $f>1$, which should always be true for a CNN.
\end{remark}

\begin{remark}[Higher dimension convolutions]
    The methods presented in this section have considered convolutions of 1D signals with a 1D filter. The same principles extend to higher dimensions, however, the zero pattern of $\bar{Z}^1$ begins to repeat in what resembles a fractal-like pattern as the dimensions increase. An example of a 2D convolution pattern is shown in Fig.~\ref{fig:2d_convolution}.
\end{remark}

\begin{corollary}
    \label{cor:input_output_sensitivity}
    The sensitivity of the network defined by (\ref{eqn_cqnntheorem}) around point $x=x_0$ is given by
    \begin{equation}
        \label{eqn:sensitivity}
        \left.\frac{\partial y}{\partial x}\right|_{x=x_0} = 2a\bar{Z}^1x_0 + b\bar{Z}^2
    \end{equation} 
\end{corollary}
\begin{proof}
    The network's output is given by
    \begin{align*}
        y &= \begin{bmatrix} x \\ 1 \end{bmatrix}^T \begin{bmatrix} a\bar{Z}^{1} & \frac{b}{2}\bar{Z}^{2} \\ \frac{b}{2}(\bar{Z}^{2})^T & c\bar{Z}^{4} \end{bmatrix} \begin{bmatrix} x \\ 1 \end{bmatrix} \\
        &= x^Ta\bar{Z}^1x + b\bar{Z}^2x + c\bar{Z}^4
    \end{align*}
    Taking the partial with respect to $x$ gives
    \begin{align*}
        \frac{\partial y}{\partial x} &= 2a\bar{Z}^1x + b\bar{Z}^2
    \end{align*}
\end{proof}

Corollary~\ref{cor:input_output_sensitivity} enables the analysis of the sensitivity of a given CQNN to changes in the input.

\section{CQNN FORMULATED AS LEAST SQUARES}
\label{sec:CQNNasLS}
The least squares formulation of the training of a CQNN follows a similar structure to the one presented for QNNs in \cite{luis2023leastsquares}. However, it takes advantage of the structure of $\bar{Z}^1$ to reduce the number of weights being solved. Consider the matrix formed by $xx^T$ using a vector $x\in\mathbb{R}^n$. As shown in Theorem~\ref{theo:CQNNasQNN}, the matrix $\bar{Z}^1$ has zeros entries for the elements corresponding to $x_ix_j$ if $|i-j|\geq f$ (see (\ref{eqn_cqnntheorem}) or Fig.~\ref{fig:CQNN}). These $x_ix_j$ combinations are all fully and exclusively contained within the first $f$ diagonals of the matrix $xx^T$, defining the diagonal numbering according to Fig.~\ref{fig:xxT_diags}. Let $\textbf{vecf}(xx^T,f)$ represent the row vector containing all the elements of the first $f$ diagonals of matrix $xx^T$
\begin{equation}
    \textbf{vecf}(xx^T,f) = \begin{bmatrix} x_1^2 \\ \svdots \\ x_n^2 \\ x_1x_2 \\ \svdots \\ x_{n-1}x_n \\ x_1x_3 \\ \svdots \\ x_{n-2}x_n \\ \svdots \\ x_1x_f \\ \svdots \\ x_{n-f+1}x_n \end{bmatrix}^T\in\mathbb{R}^{(2n-f+1)f/2 \times 1}
\end{equation}
Consider a regressor matrix of the form
\begin{equation}
    \label{eqn_LS_H}
    H=\begin{bmatrix}
        H_1+H_2 & bX
    \end{bmatrix}
\end{equation}
with
\begin{equation}
    \begin{split}
    H_1 &= \begin{bmatrix}
        \textbf{vecf}(ax_1x_1^T,f) \\ \vdots \\ \textbf{vecf}(ax_Nx_N^T,f)
    \end{bmatrix}, \quad X=\begin{bmatrix}x_1^T \\ \vdots \\ x_N^T \end{bmatrix} \\
    H_2 &= \begin{bmatrix} cU_{N,n} & O_{N,q-n} \end{bmatrix}
    \end{split}
\end{equation}
where $q=(2n-f+1)f/2$, and $O_{r,c}$ and $U_{r,c}$ are an $r\times c$ matrix of zeros and ones respectively. Using (\ref{eqn_LS_H}), problem (\ref{eqn_cqnntheorem}) with loss function $l(\hat{y},y)=||\hat{y}-y||^2$ becomes
\begin{equation}
    \begin{split}
    \text{min}\;&||\hat{y}-y||^2 \\
    \text{s.t.}\;&\hat{y}=H\theta 
    \end{split}
\end{equation}
with solution
\begin{align}
    \theta = (H^TH)^{-1}H^Ty
\end{align}
The resulting weight vector $\theta$ is of the following form
\begin{align}
    \theta = \begin{bmatrix} \bar{Z}^1_{1,1} \\ \svdots \\ \bar{Z}^1_{n,n} \\ 2\bar{Z}^1_{1,2} \\ \svdots \\ 2\bar{Z}^1_{n-1,n} \\ \svdots \\ 2\bar{Z}^1_{1,f} \\ \svdots \\ 2\bar{Z}^1_{n-f+1,n} \\ \bar{Z}^2_1 \\ \svdots \\ \bar{Z}^2_n \end{bmatrix} \in\mathbb{R}^{q+n\times1}
\end{align}

Regularization can also be introduced using regularized least squares to solve for the weight vector. The problem is given by
\begin{equation}
    \label{eqn_regLS}
    \begin{split}
        \text{min}\;&||\hat{y}-y||^2 + \beta\theta^T\theta \\
    \text{s.t.}\;&\hat{y}=H\theta 
    \end{split}
\end{equation}
with solution
\begin{align}
    \theta = (H^TH + \beta I)^{-1}H^Ty
\end{align}
where $\beta\geq0$ is a regularization term.
\begin{remark}[Regularized least squares]
    Since the regularization is being applied differently in (\ref{eqn_QNNConvex}) and (\ref{eqn_regLS}) they will only share a solution when $\beta=0$.
\end{remark}

\section{RESULTS}
\label{sec:results}
For all examples, the PyTorch library in Python was used for the BP-trained CNN networks (BP~CNN). These networks all contain 2-layers and use the ReLU activation function. The stochastic gradient descent algorithm was used with a learning rate of 0.01, a momentum factor of 0.9, and a weight decay of $10^{-5}$. All quadratic networks use $a=0.0937$, $b=0.5$, $c=0.4688$ for their activation function parameters (following the proposition from \cite{bartan2021neural} to mimic the ReLU activation function).
\subsection{System Identification}
\begin{figure}[t]
    \centering
    \includegraphics[width=0.43\textwidth]{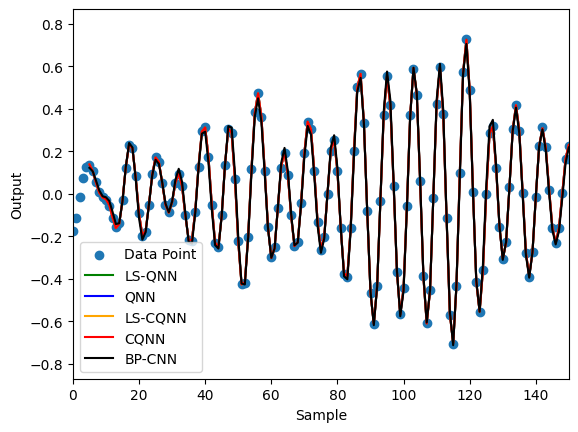}
    \caption{System identification comparison between different networks} 
    \label{fig:sysid}
\end{figure}
System identification of a flexible robot arm was performed, with the data being available from \cite{robotarmdata}. Each input sample $X_i$ and output sample $Y_i$ was constructed as follows:
\begin{subequations}
\begin{align}
        X_i &= \begin{bmatrix} u_{i-d} & \cdots & u_{i-1} & y_{i-d} & \cdots & y_{i-1} \end{bmatrix} \\
        Y_i &= \begin{bmatrix} y_{i} \end{bmatrix}
\end{align}
\end{subequations}
where $u_i$ is the arm reaction torque, $y_i$ is the arm acceleration, and $d$ is a delay term that dictates how many past samples are used to predict the next output. For this example, $d=5$ was selected, and the 1018 samples were divided using a 50/50 train/test split. All convolutional networks used a filter size of 3, and the BP~CNN used 32 filters and was trained over 10k epochs. Along with the BP~CNN, the compared methods were the QNN and CQNN convex training methods proposed in \cite{bartan2021neural}, the least squares QNN (LS-QNN) proposed in \cite{luis2023leastsquares}, and the least squares CQNN (LS-CQNN) from this paper. The results are plotted in Fig.~\ref{fig:sysid} with the training mean squared error (MSE), testing MSE, and training time shown in Table~\ref{tab:sysid}. The CQNN networks outperformed both the QNN and the BP~CNN in terms of testing MSE, with the LS-CQNN method having the lowest training time. Furthermore, compared to a BP-trained network, the LS-CQNN method provides an analytic quadratic model of the system. Using this quadratic model and equation (\ref{eqn:sensitivity}), the maximum sensitivity of the output to each input is found as
\begin{align*}
        \frac{\partial y_i}{\partial u_{i-d}} &= 
        \begin{bmatrix}
        0.68 & 1.56 & 2.02 & 1.59 & 0.51
        \end{bmatrix}^T \\
        \frac{\partial y_i}{\partial y_{i-d}} &= 
        \begin{bmatrix}
        2.65 & 5.70 & 5.64 & 2.87 & 3.52
        \end{bmatrix}^T
\end{align*}
This shows that the system is more sensitive to previous outputs compared to inputs, and that the output is affected mostly by the input after a few delays, suggesting a possible lag between the input and the reaction of the system.

\begin{table}[b]
        \centering
        \caption{System identification training/testing MSE and training time for various neural network solutions}
        \begin{tabular}{| c | c c c |}
        \hline
        Network & Training MSE & Testing MSE & Train Time [s]\\ 
        \hline
        QNN \cite{bartan2021neural} & $7.26\times10^{-6}$ & $1.14\times10^{-5}$ & 1.228 \\ 
        LS-QNN \cite{luis2023leastsquares} & $7.21\times10^{-6}$ & $1.14\times10^{-5}$ & 0.040 \\ 
        CQNN \cite{bartan2021neural} & $7.99\times10^{-6}$ & $1.01\times10^{-5}$ & 5.917 \\ 
        LS-CQNN & $7.99\times10^{-6}$ & $1.01\times10^{-5}$ & 0.019 \\ 
        BP-CNN & $4.54\times10^{-4}$ & $4.90\times10^{-4}$ & 9.092 \\ 
        \hline
        \end{tabular}
        \label{tab:sysid}
\end{table}

\subsection{GPS Signal Emulation}

\begin{figure}[t] 
        \centering
        \includegraphics[width=0.4\textwidth]{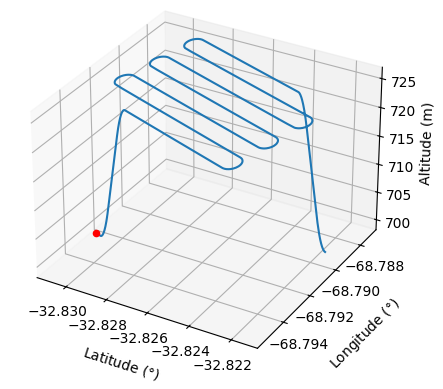}
        \caption{Synthetic drone trajectory with starting position indicated in red} 
        \label{fig:navego_traj}
\end{figure}

\begin{figure}[t] 
        \centering
        \includegraphics[width=0.48\textwidth]{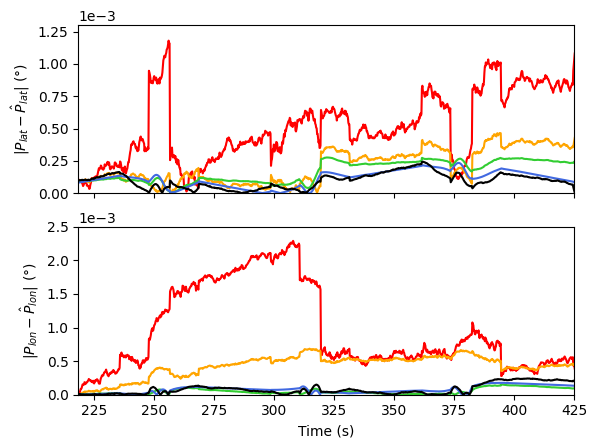}
        \caption{Comparison of prediction errors over testing set for BP~CNN (black) and LS-CQNN with various regularization terms $\beta$ (red=0.1, orange=1, green=10, blue=100) for GPS latitude and longitude predictions} 
        \label{fig:gps_results}
\end{figure}
Typical outdoor drone navigation uses both an inertial measurement unit (IMU) and GPS data for sensor fusion to produce accurate position and attitude estimations. GPS outages present an issue for this type of sensor fusion, because when GPS measurements are not present, the estimation errors grow rapidly. Recent work has proposed the training of neural networks to mimic the GPS signal during periods of outages \cite{gpsDenied_2017,deepGPSDenied_2022,attentionGPSDenied_2023}. For this example, the CQNN will be used on synthetic drone flight data created from the NaveGo toolbox for MATLAB \cite{navego1, navego2}. The trajectory is shown in Fig.~\ref{fig:navego_traj}. Following the model proposed in \cite{gpsDenied_2017}, the input and output vectors have the following form
\begin{subequations}
\begin{align}
        X_i &= 
        \begingroup 
        \setlength\arraycolsep{2pt}
        \begin{bmatrix}
        a_x^T & a_y^T & a_z^T & \omega_x^T & \omega_y^T & \omega_z^T & \theta_x^T & \theta_y^T & \theta_z^T
        \end{bmatrix}^T \endgroup \\ 
        Y_i &= \begin{bmatrix}
        \Delta P_{lat} & \Delta P_{lon}
        \end{bmatrix}^T
\end{align}
\end{subequations}
where $a_x,a_y,a_z\in\mathbb{R}^{r}$, $\omega_x,\omega_y,\omega_z\in\mathbb{R}^{r}$, and $\theta_x,\theta_y,\theta_z\in\mathbb{R}^{r}$ are the IMU acceleration, IMU angular rate, and inertial navigation system (INS) attitude in the $x,y,z$ axes of the body frame of the the drone and $r$ is the sampling rate ratio between the IMU and the GPS. Furthermore, $\Delta P_{lat}, \Delta P_{lon}\in\mathbb{R}$ are the change in latitude and longitude of the drone in the navigation frame. The GPS operates at 5 Hz, whereas the IMU and INS operate at 200 Hz, leading to $r=40$ for this example. The data contains a total of 2186 samples, which are divided using a 50/50 train/test split. A BP~CNN was trained on the data over 2000 epochs. A least squares CQNN (LS-CQNN) with a filter size of 7 was trained using regularized least squares with various $\beta$. The BP~CNN took 13.11 seconds, whereas the LS-CQNN took 4.67 seconds on average. The results for latitude and longitude predictions showing the test set are shown in Fig.~\ref{fig:gps_results}. In addition to a faster training time, the LS-CQNN would also permit a sensitivity analysis of the model.

\section{CONCLUSIONS}
\label{sec:conclusion}
This paper proposes a formulation of the training of a quadratic convolutional neural network as a least squares problem. This method has the benefit of providing both an analytic expression for the globally optimal weights and a quadratic input-output relationship. These allow for a more complete analysis of the network in the context of system theory when compared to conventional convolutional networks. Furthermore, compared to the CQNN training method presented in \cite{bartan2021neural}, the least squares formulation results in lower training times via a reduction in the total number of weights. The results were applied to both a system identification problem and a GPS signal emulation problem.

\bibliographystyle{IEEEtran}
\bibliography{IEEEabrv,biblio}
\end{document}